\pdfoutput=1

\documentclass[11pt]{article}

\usepackage[]{acl}

\usepackage{times}
\usepackage{latexsym}

\usepackage[T1]{fontenc}

\usepackage[utf8]{inputenc}

\usepackage{microtype}

\usepackage{times}
\usepackage{latexsym}

\usepackage{algorithm,amsmath}
\usepackage{pifont}
\usepackage{bbding}
\usepackage{amsfonts}
\usepackage{mathrsfs}
\usepackage[noend]{algorithmic}
\usepackage{url}
\usepackage{makecell}
\usepackage{graphicx}
\usepackage{microtype}
\usepackage{xcolor}
\usepackage{csquotes}
\usepackage{booktabs}
\usepackage{caption}
\usepackage{subcaption}
\usepackage[flushleft]{threeparttable}

\title{Locally Aggregated Feature Attribution on Natural Language Model Understanding}

\author{$\textbf{Sheng Zhang}^{1}$\ \ $\textbf{Jin Wang}^{2}$  \ \ $\textbf{Haitao Jiang}^{3}$  \ \  $\textbf{Rui Song}^{2, 3}$\\
  $\textbf{ }^1$AWS AI Labs \\
  $\textbf{ }^2$Amazon Core AI\\
  $\textbf{ }^3$Department of Statistics,
  North Carolina State University\\
  \texttt{\{zshe, jiwngn\}@amazon.com}, \ \ \texttt{\{hjiang24, rsong\}@ncsu.edu}  \\
  }

\begin{document}
\maketitle
\begin{abstract}
With the growing popularity of deep-learning models, model understanding becomes more important. Much effort has been devoted to demystify deep neural networks for better interpretability. Some feature attribution methods have shown promising results in computer vision, especially the gradient-based methods where effectively smoothing the gradients with reference data is key to a robust and faithful result. However, direct application of these gradient-based methods to NLP tasks is not trivial due to the fact that the input consists of discrete tokens and the ``reference'' tokens are not explicitly defined. In this work, we propose Locally Aggregated Feature Attribution (LAFA), a novel gradient-based feature attribution method for NLP models. Instead of relying on obscure reference tokens, it smooths gradients by aggregating similar reference texts derived from language model embeddings. 
For evaluation purpose, we also design experiments on different NLP tasks including Entity Recognition and Sentiment Analysis on public datasets as well as key feature detection on a constructed Amazon catalogue dataset.
The superior performance of the proposed method is demonstrated through experiments. 
\end{abstract}

\section{Introduction}

With the growing popularity of deep-learning models, model understanding becomes more and more critical in many folds.
In one aspect, model understanding helps us understand what the model is doing by identifying crucial features among unstructured raw data. For example, \citealp{shrikumar2017learning} utilized the model explainability technique to discover motifs in regulatory DNA elements from distinct molecular signatures in the field of Genomics.
In another aspect, model understanding helps people audit or debug the deep models. An interesting example is that Ribeiro et al. \citep{ribeiro2016should} found that their image classification model sometimes misclassifies a husky as a wolf. The model explainability tool reveals that their model relies on the snow in the background rather than the appearance when distinguishing the two animals.
More importantly, model understanding helps gain trust when making important decisions based on the model. In the NLP domain, deep language models are quickly evolving and show superior performance in various benchmark tasks. However, even experts struggle to understand the mechanism of complex language models.

Much effort has been devoted to demystifying the ``black box'' of deep models. A natural idea is through feature attribution, explaining the model by attributing the prediction to each input feature according to how much it affects the model output, of which two main directions emerge. 
One is model agnostic approaches including Shapley regression values \citep{shapley1953value} and LIME \citep{ribeiro2016should}.
We can apply these methods regardless of the model structure, however, they could suffer from computational inefficiency in the scenario of high dimensional input space and complex deep models when making inferences across all possible permutations or with small perturbations in the local neighborhood. 

Another direction is model-specific approaches which look into the internal model mechanism to understand specific models.
Gradient-based feature attribution models are often adopted to explain neural networks since gradients can be easily accessed through back-propagation, which gives a great computational advantage over model-agnostic methods. Since the gradient map itself is often noisy and challenging to interpret, most gradient-based methods aim to stabilize the feature attribution score by smoothing the gradients or learning from the reference data \citep{sundararajan2017axiomatic, smilkov2017smoothgrad, lundberg2017unified}. However, direct application of these gradient-based methods to NLP problems is not trivial, due to the fact that the input consists of discrete tokens and the ``reference'' tokens are not explicitly defined.

In this paper, we propose \textbf{L}ocally \textbf{A}ggregated \textbf{F}eature \textbf{A}ttribution (\textbf{LAFA}), a novel gradient-based approach that leverages sentence-level embedding as a smoothing space for the gradients, motivated by the observation that the feature attribution is often shared by similar text inputs. For example, key features in product descriptions on an online marketplace are often shared by similar products. We implement a neighbor-searching method to ensure the quality of neighboring sentences. 

Furthermore, to evaluate feature attribution methods in NLP, we consider two situations. For datasets with golden labels of feature score, we use the Area Under Curve (AUC) or Pearson correlation as the performance metric. 
As for datasets without golden labels, we conduct a similar evaluation task following prior works \cite{shrikumar2017learning, lundberg2017unified} by masking tokens with high importance scores and find the change in the predicted log-odds.

In summary, our contributions are threefold: 
First, we build a novel context-level smooth gradient approach for feature attribution in NLP. 
The key ingredients of our method are constructing an appropriate aggregation function over the smoothing space. 
Second, to the best of our knowledge, this is the first proposal to conduct numerical studies on multiple NLP tasks, including Entity Recognition and Sentiment Analysis, for feature attribution.
Third, our method achieves superior performance compared with the state-of-the-art feature attribution methods.

The paper is organized as follows. 
Section \ref{sec:background} elaborates the current challenges of feature attribution in NLP and recaps the preliminaries about gradient-based feature attribution approaches.
The proposed feature attribution method is described in section \ref{sec:framework}, followed by a review of other existing approaches in Section \ref{sec:related_work}.
The evaluation tasks and the application results on NLP are presented in Section \ref{sec:exp}. 

\begin{figure*}
  \centering
  \includegraphics[width=1\textwidth]{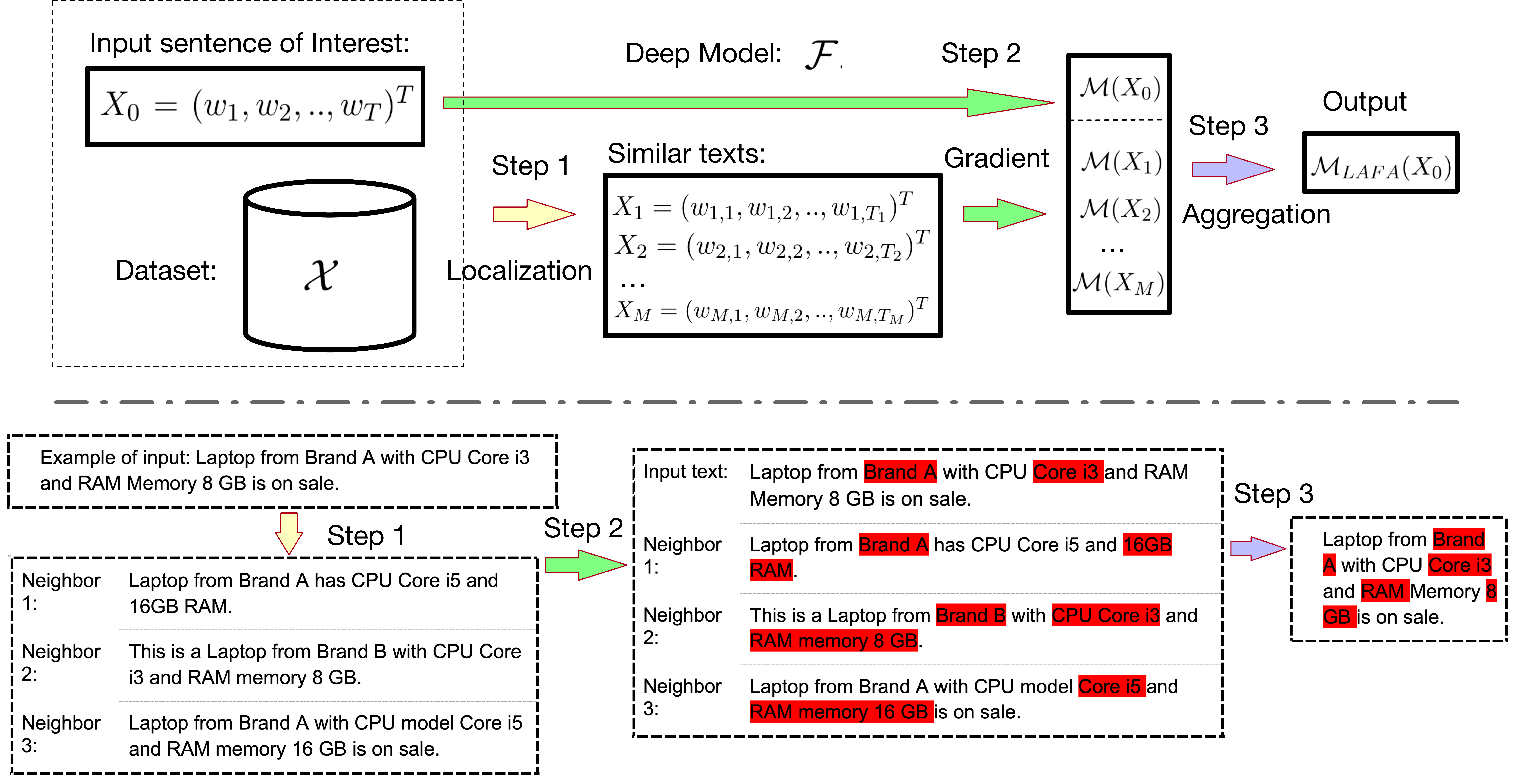}
 \caption{
 \textit{Upper panel} shows the overview of LAFA methods;
 \textit{Lower panel} provides a motivating example of LAFA method. In this motivating example, the input text is a description of computer. The key features of the computer should include ``Brand'', ``CPU type'' and ``RAM size''.  
 The simple gradient method may miss certain feature, such as  ``RAM size'' in the example, while the gradients on similar texts can provide more contexts. The proposed method is constructed to aggregate the information from similar texts summarized in Algorithm \ref{algotable}.
 }
 \label{fig:overview}
\end{figure*}

\section{Feature Attribution in NLP}\label{sec:background}
\paragraph{Challenge}
Direct application of gradient-based methods to NLP problems is not trivia. 
There are three main challenges. 
First, NLP models consist of non-differentiable discrete input tokens, hence the gradient hook can only reach out to the embedding space and gradient-based feature attribution methods are not directly applicable to word tokens. 

Second, the reference data in NLP are difficult to define. It is studied by Sundararajan. et al \citep{sundararajan2017axiomatic} that using the gradient as the feature attribution may suffer from the problems of \textit{model saturation} or \textit{thresholding}.
Model saturation means the perturbation of some elements in the input cannot change the output, and the thresholding problem indicates discontinuous gradients can produce misleading importance scores.  
Such problems can be addressed by comparing the difference between the gradient of input and reference data.
The guiding principle to select reference data is to ask ourselves that ``what am I interested in measuring differences against?''
For example, in the tasks of binary classification on DNA sequence inputs, the reference data are chosen as the expected frequencies of DNA sequence or randomly shuffling the original sequence. 
However, in NLP tasks, randomly shuffling texts as reference may not be grammatically sensible.

Lastly, we note that the evaluation of the language model is much more challenging than the explanations of the images. In the image application, the important features of an image obtained from feature attribution methods can be visually validated by checking the composition of objects. 
However, the detected important features in language may require more domain knowledge to validate. 

\paragraph{Problem Definition}
Feature attribution task can be formally formulated as follows.
A deep model $\mathcal{F}$ is provided to be explained, which is fine-tuned on dataset $\mathcal{X}$.
The input sentence is denoted as $X_0 = (w_1, w_2, .., w_T)^T$ where $w_i$ represents $i$-th word.
The goal for feature attribution is to determine function $\mathcal{M}(\cdot)$ by quantifying the importance score of each word $\mathcal{M}(x) = (m_1, m_2, .., m_T)^T$, where $m_i$ denotes the importance score for $w_i$.


\paragraph{Simple Gradient as Feature Attribution}
As illustrated in the first challenge above, in NLP models, directly taking derivative on each word is infeasible due to the non-differentiable embedding layer. 
We can resolve the challenge as follows.

The fist layer of the NLP model usually maps input discrete tokens to embedding from a pre-defined dictionary. 
\begin{equation}
    h_{0, i} = emb(w_i),\ i = 1,2,..,T,
    \label{eq: w2v}
\end{equation}
where $h_{0, i} \in \mathbb{R}^d$ represents the word embedding for $w_i$. 
This step is non-derivative. But we can obtain the derivative of output with respect to the word embedding:
\begin{equation}\label{eq:simple_gradient_S}
    \mathcal{S}(H_0) = \partial{\mathcal{F}}/ \partial{H_0} \in \mathbb{R}^{T \times d},
\end{equation}
where $H_0 = (h_{0, 1}, h_{0, 2}, .., h_{0, T})^T \in \mathbb{R}^{T \times d}.$
Then, we consider the feature attribution score of a token $\mathcal{M}(X) \in \mathbb{R}^T$ as the sum of squares of the gradients with regard to each word embedding dimension:
\begin{equation}\label{eq:simple_gradient}
    \mathcal{M}(X)_i = \sum_{j = 1}^d\mathcal{S}(H_0)_{i,j}^2, \ i = 1,2,.., T.
\end{equation}

However, simply using the gradients of one token as feature attribution would lead to noisy results \cite{sundararajan2017axiomatic}. The next section describes a novel feature attribution approach that smoothes the gradients by leveraging similar input texts.

\section{The Proposed Framework: LAFA} \label{sec:framework}

The proposed method contains three steps: (1)~find the appropriate neighbors of the input text for gradient smoothing; (2)~calculate the gradients of texts as well as neighbors;
(3)~aggregation of the gradients.
The proposed framework is summarized in the upper panel of Figure \ref{fig:overview}.
One motivating example is shown in the lower panel of Figure \ref{fig:overview}.
In this motivating example, the input text is a description of computer. The key features of the computer should include ``Brand'', ``CPU type'' and ``RAM size''. 
The simple gradient method may miss certain feature, such as  ``RAM size'' in the example, while the gradients on similar texts can provide more contexts.
The proposed method is constructed to aggregate the information from similar texts.

\paragraph{Step I: Context-level Localization}
Given the input text $X_0 \in \mathcal{X}$, where $\mathcal{X}$ denotes the input datasets, the goal is to find similar texts $\mathcal{X}_{sim} = \{X_1,X_2, .., X_M\} \subset \mathcal{X}$ such that the feature attributions of $X_0$ and $X_j \in \mathcal{X}_{sim}$ are similar under a defined similarity metric. 

To obtain similar texts $\mathcal{X}_{sim}$, we first define an encoder that maps the text with discrete word tokens to a continuous embedding vector; then, in the embedding space, similar texts are found in the neighbor of $X_0$.
To be specific, let $\mathcal{H}_{encoder}$ denote the mapping from input to one of the hidden layers in deep model $\mathcal{F}$. 
$\mathcal{X}_{sim}$ can be obtained by choosing closest texts in the dataset as follows:
\begin{equation}\label{eq:similar_text}
    \begin{split}
        & X \in \mathcal{X}  \\
        s.t. \ &||\mathcal{H}_{encoder}(X) - \mathcal{H}_{encoder}(X_0)||_2 < \epsilon
    \end{split}
\end{equation}
where $||\cdot||_2$ represents $L^2$ norm. $\epsilon$ is a threshold score to guarantee that founded neighbors are similar to the center text $X_0$ to improve the faithfulness of aggregation. 
In our application, a fixed quantile served as the cut-off rate of L2 distance for all possible pairs is chosen as the threshold score to filter the nearest-neighbor result.
During inference time, we apply the hidden layer encoder $\mathcal{H}_{encoder}$ to all the input datasets and index, then using FAISS \footnote{\url{https://github.com/facebookresearch/faiss} (MIT license)} \citep{JDH17} offline.
FAISS is an efficient, open-source library for similarity search and clustering on dense vectors, which can be applied to large-scale vectors.

The output of this step, $\mathcal{X}_{sim}$ can be viewed as the reference data to smooth the feature attribution of $X_0$, which addresses the second challenge listed in Section \ref{sec:background}. 




\paragraph{Step II: Taking Gradients}\label{sec:step2} 
According to Equation \eqref{eq:simple_gradient}, the gradient of $X_i$ can be denoted as $\mathcal{M}(X_i) := (m_{i,0}, m_{i,1}, .., m_{i,T_i})^T$ for $i = 0, 1, .., ,M$ where $T_i$ represent the token length of $X_i$.
To be noticed that our proposed method can be easily extended to variants of simple gradient including smooth gradient or integrated gradient methods \cite{smilkov2017smoothgrad, sundararajan2017axiomatic} in Step II.  

\paragraph{Step III: Aggregation over Multiple Feature Attribution}\label{sec:step3} 
Our goal is to smooth the gradient $\mathcal{M}(X_0)$ by aggregating the gradients of similar text inputs:
\begin{equation}\label{eq:smooth_gradient_our}
    \begin{split}
        \mathcal{M}_{LAFA}(X_0) = & AGGREGATE(\mathcal{M}(X_0);\\  & \mathcal{M}(X_1), .., \mathcal{M}(X_M))
    \end{split}
\end{equation}

Since the lengths of $X_0$, $X_1$,..,$X_M$ may vary, the lengths of gradients $\mathcal{M}(X_0),$ $\mathcal{M}(X_1)$,..,$\mathcal{M}(X_M)$ are different as well. Consequently, aggregation by simply taking the average is infeasible. 
Following the intuition that, the tokens with high gradients in $\mathcal{X}_{sim}$ should be important in $X_0$, we propose the following aggregation function:
\begin{equation}\label{eq:smooth_gradient_our2}
    \begin{split}
        \mathcal{M}_{LAFA}(X_0) & =\mathcal{M}(X_0) 
        + \lambda ( \mathcal{E}(w_{0, 1};  \mathcal{X}_{sim}), .., \\
        & \mathcal{E}( w_{0, T } ; \mathcal{X}_{sim}))^T,
    \end{split}
\end{equation}
where $\lambda$ is a hyper-parameter for leveraging the feature attribution from similar inputs. 
$\mathcal{E}(w; \mathcal{X}_{sim}) $ is a scalar representing the importance of token $w$ obtained from the neighbor inputs $\mathcal{X}_{sim}$.
Formally, it can be defined as 
\begin{equation}
    \mathcal{E}(w; \mathcal{X}_{sim}) = \frac{1}{|\mathcal{X}_{sim}|}\sum_{i = 1}^{|\mathcal{X}_{sim}|}\sum_{k = 1}^{T_i} \frac{m_{i,k} \times k(h, h_{i, k})}{T_i},
    \label{eq:kernel_form}
\end{equation}
where $h$, $h_{i, k}$ are the word embedding of $w$ and $w_{i, k}$ as in Equation \eqref{eq: w2v} respectively, and $k(\cdot, \cdot)$ is a kernel function \citep{hofmann2008kernel} (examples of kernel function are listed in the Appendix \ref{appendix:C}. ).
According to Equation \eqref{eq:kernel_form}, if word $w$ and $w_{i, k}$ have a high similarity, then inner product between the embeddings $h$ and $h_{i, k}$ in the kernel space would be large, which would assign a large weight to the corresponding importance score $m_{i,k}$. 
On the contrary, dissimilar word $w_{i, k}$ in $\mathcal{X}_{sim}$ has little effect to the word $w$ in $\mathcal{E}(w; \mathcal{X}_{sim})$.
The whole process is summarized in Algorithm \ref{algotable}.

\begin{algorithm}[H]
      \caption{Feature attribution method with smoothing over similar inputs.}
  \begin{algorithmic}[1]
  \STATE \textbf{Input:} Text of interest $X_0$, input datasets $\mathcal{X}$, and fine-tuned deep model $\mathcal{F}$. 
  \STATE \textbf{Output:} Feature attribution of $X_0$
  \STATE \textbf{Step I: Localization} 
  \STATE Construct encoder $\mathcal{H}$ which maps from input space to the space of hidden layer in $\mathcal{F}$.
  \STATE Obtain the similar texts set $\mathcal{X}_{sim} = \{X_1, X_2,..,X_M\}$ of $X_0$ according to Equation \eqref{eq:similar_text}.
  \STATE \textbf{Step II: Taking Gradient}
  \STATE Calculate the gradient of texts $X_0, X_1, .., X_M$ according to Equation \eqref{eq:simple_gradient}.
  \STATE \textbf{Step III: Aggregation}
  \STATE Smooth the gradient of text of interest over the gradient of similar texts according to Equation \eqref{eq:smooth_gradient_our2}. 
  \STATE Output the aggregated gradient $\mathcal{M}_{LAFA}(X_0)$ as the feature attribution.
  \end{algorithmic}
      \label{algotable}
\end{algorithm}

\paragraph{Discussion of Faithfulness}
One important criteria for model explainability method is ``faithfulness'', which refers to how accurately it reflects the true reasoning process of the model \cite{jacovi2020towards}.
In our proposed method, the original input $X_0$ is infused with similar texts in the input dataset $\mathcal{X}$ for better interpretation. 
Since the deep model $\mathcal{F}$ is also trained on $\mathcal{X}$, using similar texts $\mathcal{X}_{sim} \subset \mathcal{X}$ to facilitate explanation will not violate the faithfulness.

In the localization step (Step I), out of the consideration about faithfulness, we do not use popular bi-encoder frameworks, such as S-BERT \cite{reimers2019sentence} or DenseRetrival \cite{karpukhin2020dense}, to obtain similar neighbors. Because it will involve an extra black box model when explaining deep model $\mathcal{F}$.

\section{Related Work} \label{sec:related_work}

In NLP, transformer-based models yield great successfulness and some works focus on explaining the attention mechanism. For example, \citealp{serrano2019attention} and \citealp{jain2019attention} inspected a single attention layer and found out that attention weights only weakly and inconsistently correspond to feature importance;
\citealp{wiegreffe2019attention} argued that we cannot separate the attention layer and should view the entire model as a whole.
In this section, We mainly review the gradient-based methods for feature attribution.

\paragraph{Feature Attribution on Single Input}
Simonyan et al \citep{simonyan2013deep} computed the ``saliency map'' denoted as \textit{Simple Gradient} from the derivative of the output with respect to the input in an image classification task.
In the NLP application, ``saliency map'' is obtained as the derivative of the output with respect to the word embedding as in Equation \eqref{eq:simple_gradient_S}.
However, ``saliency map'' can be visually noisy. Several methods are proposed to improve the gradient method from different perspectives.
\textit{Gradient*Input} method \citep{shrikumar2017learning} improves the
visual sharpness of the ``saliency map'' by multiplying gradient with the input itself.
In NLP, we can write it as:
    \begin{equation*}\label{eq:grad*input}
    \begin{split}
        \mathcal{S}_{Grad*Input}(H_0) & = H_0 \times \mathcal{S}(H_0) \\
        \mathcal{M}_{Grad*Input}(X)_i & = \sum_{j = 1}^d\mathcal{S}_{Grad*Input}(H_0)_{i,j}^2 .
    \end{split}
    \end{equation*}

\textit{Layerwise Relevance Propagation} method \citep{bach2015pixel} is shown to be equivalent to the \textit{Gradient*Input} method up to a scaling factor.
\textit{Smooth Gradient} method \citep{smilkov2017smoothgrad} smoothes the feature attribution score by adding random noises to the input and taking average of the gradients from noisy inputs, formally:
    \begin{equation*} \label{eq:smooth_grad}
        \begin{split}
            \mathcal{S}_{SmoothGrad}(H_0)
          & \approx \frac{1}{N}\sum_{k = 1}^{N}\mathcal{S}(H_0 + \epsilon_k),  \\ 
          & \epsilon_k \sim N(0, \sigma^2), \\
          \mathcal{M}_{SmoothGrad}(X)_i & = \sum_{j = 1}^d\mathcal{S}_{SmoothGrad}(H_0)_{i,j}^2 .
        \end{split}
    \end{equation*}
\textit{Guided Backpropagation} method \citep{springenberg2014striving} modifies the back-propagation to preserve negative gradients in the ReLU activation layer which also sharpens the ``saliency map'' visually. Other methods, such as \textit{Grad-CAM} or \textit{Guided-CAM} \citep{selvaraju2017grad}, are applicable to specific architecture of neural networks in the field of computer vision. 

Since language models like BERT do not contain specific architecture utilized in Guided Back-propagation or Grad-CAM method, we ignore the mathematical formulation here.

\paragraph{Feature Attribution on Input with Reference Data}

\textit{Integrated Gradient} method computes the feature score by integrating the gradients from single pre-determined reference input to the target input \citep{sundararajan2017axiomatic}. In computer vision problems, black image is usually considered as the reference data, and integrating gradients from the black image to the input image represents the feature attribution of the input image.
In NLP problems, we can define the $i$-th element of feature attribution as:
\begin{equation*} \label{eq:integrad}
\begin{split}
    & \mathcal{S}_{InteGrad}(H_0)_{ij} 
    \\ & \approx \frac{H_{0, ij} - H_{ij}'}{N} \sum_{k = 1}^N \mathcal{S}( H' + k \frac{H_0 - H'}{N})_{ij}, \\
    &  \mathcal{M}_{InteGrad}(X)_i = \sum_{j = 1}^d\mathcal{S}_{InteGrad}(H_0)_{i,j}^2 .
\end{split}
\end{equation*}
where $H'$ denotes the embedding of reference text.

\textit{SHAP-Gradient} method which combines ideas from Integrated Gradient and Smooth Gradient into a single expected value equation \citep{lundberg2017unified} .
To be specific, the feature attribution is defined from:
\begin{equation*}\label{eq:shap_grad}
    \begin{split}
        & \mathcal{S}_{ShapGrad}(H_0) 
         \approx \frac{1}{N} \sum_{k = 1}^N\mathcal{S}(\alpha_k H_0 + (1 - \alpha_k)H_k), \\
        & \mathcal{M}_{ShapGrad}(X)_i  = \sum_{j = 1}^d\mathcal{S}_{ShapGrad}(H_0)_{i,j}^2 .
    \end{split}
\end{equation*}
where $\alpha_k \sim U(0, 1)$ denotes uniform distribution from zero to one, $H_k \in \mathcal{H}_{ref}$ denotes the embedding of reference text.

\textit{DeepLIFT} \citep{shrikumar2017learning} assigns the feature score by comparing the difference of contribution between input and some reference inputs via gradient. 
As discussed in \citep{lundberg2017unified}, DeepLIFT can be considered as an approximation of Shapley Value estimation. 
Specifically, as in the application of SHAP \footnote{\url{https://github.com/slundberg/shap} (MIT License)}, the feature attribution of \textit{SHAP-Deep} as a variant of \textit{DeepLIFT} is defined as:
    \begin{equation*}\label{eq:shap_deep}
    \begin{split}
        & \mathcal{S}_{ShapDeep}(H_0)
         \approx \frac{1}{N}\sum_{k = 1}^N \mathcal{S}(H_k) \times (H_0 - H_k). \\
        & \mathcal{M}_{ShapDeep}(X)_i  = \sum_{j = 1}^d\mathcal{S}_{ShapDeep}(H_0)_{i,j}^2 .
    \end{split}
    \end{equation*}

\section{Experiments}\label{sec:exp}

In this section, we compare the proposed method to the state-of-the-art feature attribution methods under different use cases.

\subsection{Case I: Feature Attribution on Relation Classification Model}\label{subsection : case_one}

\begin{table}[H]
\centering
\resizebox{0.8\columnwidth}{!}{%
\begin{tabular}{lccc}
\hline
Dataset & Precison & Recall & F1   \\ \hline
NYT10     & 94.8     & 93.3   & 94.1     \\ 
Webnlg  & 93.6     & 82.5   & 87.7      \\ \hline   
\end{tabular}
}
\caption{Fine-tuned result on multi-label relation classification task.}
\label{table:use_case_1_re}
\end{table}

\paragraph{Motivation} Relation Classification is beneficial to downstream problems, including question answering and knowledge graph (KG) construction tasks \citep{wen2016network, dhingra2016towards, dong2020autoknow}. 
With the development of deep language model, existing relation extraction methods have achieved significant performance in relation classification task \citep{soares2019matching, wei2019novel}. 
We hope to better understand the features in the text that help deep language model to classify the relations. 
In this use case, we fine-tune a deep language model with relation as labels.
With the fine-tuned model, the feature attribution technique is applied to identify the entities in the text as important features.

\paragraph{Data}
We use the public available datasets NYT10 \citep{riedel2010modeling} and Webnlg \citep{gardent2017creating} for numerical study.
\citealp{zeng2018extracting} adapted the original dataset for relation extraction task.
We follow the same setting as in \citealp{zeng2018extracting}, i.e. NYT10 dataset contains 56,196/5,000/5,000 plain texts in train/val/test set, 24 relation type, averaged 2.01 relational triples in each text. Webnlg dataset contains 5,019/500/703 plain texts in train/val/test set, 211 relation type, averaged 2.78 relation triples in each text. 

\paragraph{Language Model}

We fine-tuned BERT-base models to classify the relations for NYT10 and Webnlg datasets, respectively. 
We use the plain text as input $X$, and relations as multi-class label $Y$ in the model fine-tuning.
Since multiple relations may exist in single text, we use the \textit{Sigmoid} activation in the output layer.
Mean Square Error (MSE) is used as loss objective and Adam \citep{kingma2014adam} is adopted as the optimizer.
The micro Precision, Recall and F1 results are reported in Table \ref{table:use_case_1_re} with 0.5 threshold of output score. 
From the result, the F1 scores are high for both NYT10 and Webnlg dataset, hence we can apply feature attribution methods to the fine-tuned models and identify the important features in the text which help to classify the relations.

\begin{table}[H]
\resizebox{\columnwidth}{!}{
\begin{tabular}{l|cc}
\hline
                 & \multicolumn{2}{c}{AUC}        \\ \cline{2-3} 
Method           & NYT           & Webnlg         \\ \hline
Rand             & 0.498 (0.143) & 0.501 (0.121)  \\
SimpleGrad       & 0.949 (0.071) & 0.670 (0.135)  \\
InputGrad        & 0.953 (0.061) & \textbf{0.713} (0.120)  \\
InteGrad         & 0.948 (0.077) & 0.663 (0.126)  \\
SmoothGrad       & \textbf{0.960} (0.064) & 0.664 (0.142)  \\
SHAP + Zero      & 0.805 (0.213) & 0.670 (0.133)  \\
SHAP + Ref. & 0.872 (0.169) & 0.675 (0.133)  \\
\hline
LAFA             & \textbf{0.958} (0.060)  & \textbf{0.724} (0.115)  \\ \hline
\end{tabular}
}
\caption{Feature attribution result on Relation Classification model (Case I). Top two results are highlighted in bold.}
\label{table:use_case_1_main}
\end{table}

\paragraph{Evaluation Metric}

In datasets, NYT10 and Webnlg, the positions of entities in triples are provided.
Therefore, we can constructed the golden feature attribution label as follow.
For text $X = (w_1, w_2,..,w_T)^T$ and triple $(s, r, o)$, where subject $s = (w_{i}, .., w_{j})$ and object $o = (w_{k}, .., w_{s})$ are words shown in the text from positions $i$ to $j$ and $k$ to $s$, respectively. 
The gold labels of feature attribution for relation $r$ is constructed as $$\mathcal{M}_{gold}(X) = (0, .. , 0, 1, .. ,1, 0,... , 0, 1, .. ,1, .. 0 )^T$$ where we set 1 from positions $i$ to $j$ as well as $k$ to $s$ and set 0 on other positions.

We use the evaluation metric Area under Curve (AUC) to compare the feature attribution $\mathcal{M}_{\cdot}(X)$ and $\mathcal{M}_{gold}(X)$ for the test dataset. 
AUC ranges from 0 to 1, higher AUC represents the the feature attribution result is closer to the gold feature attribution.


\paragraph{Main Results}

The results of AUC under different methods are summarized in Table \ref{table:use_case_1_main}. 
The popular feature attribution methods are listed and compared. 
More introduction about the competitors can be found in Section \ref{sec:related_work}. 
``Rand'', as a baseline method, denotes that the feature score is randomly assigned, therefore, the AUC score is about 0.5.
InputGrad method performs better than the SimpleGrad, showing the effeteness of Taylor approximation of layer-wise relevance propagation. 
``SHAP + Zero'' means zero references are used in SHAP and ``SHAP + Ref.'' means $\mathcal{X}_{sim}$ is used as references. SHAP-based methods show low AUC values, because such methods aggregate the gradients of input and reference by simply taking average aggregation (see details in Section \ref{sec:related_work}), which is not meaningful in NLP tasks.
From the result, our method LAFA achieves a superior performance in Webnlg dataset and comparable performance in NYT dataset, which indicates that our feature attribution method can identify entities well.

\subsection{Case II: Feature Attribution on Sentiment Analysis}

\begin{table}
 \resizebox{\columnwidth}{!}{
\begin{tabular}{l|cc}
\hline
                 & \multicolumn{2}{c}{Pearson Correlation}  \\ \cline{2-3} 
Method           &   SST-2         &  SST     \\ \hline
Rand             &   0.039(0.074)          & 0.040(0.072)   \\
SimpleGrad       &   0.441(0.083)          & 0.430(0.081)    \\
InputGrad        &   0.456(0.080)          & 0.448(0.078)   \\
InteGrad         &   0.468(0.071)          & 0.454(0.072)    \\
SmoothGrad       &   \textbf{0.484(0.073)} & \textbf{0.471(0.073)}     \\
SHAP + Zero       &   0.400(0.087)         & 0.392(0.085)      \\
SHAP + Ref.       &   0.279(0.093)         & 0.278(0.091)       \\
 \hline
LAFA             &   \textbf{0.494(0.070)} & \textbf{0.481(0.070)}      \\  \hline
\end{tabular}
}
    \caption{Feature attribution result on Sentiment Analysis Model (Case II).}
\label{table:use_case_2_main}
\end{table}

\paragraph{Motivation} 
The goal of the sentiment classification task is to classify a text into a sentiment categories such as positive or negative sentiment \citep{aghajanyan2021muppet-sst2, raffel2019exploring-sst2, jiang-etal-2020-smart-sst2}.
In this use case, we hope to explain the deep sentiment classification model and obtain sentiment factors that drive the model to identify the sentiment. 

\paragraph{Data}
The Stanford Sentiment Treebank (SST) \citep{socher2013recursive} is a sentiment analysis dataset collected from English movie reviews 
\citep{pang2005seeing}. For all $9,645$ sentences in SST, Amazon Mechanical Turk labeled the sentiment for words/phrases/sentences yielded from the Stanford Parser \citep{manning2014stanford} on a scale between $1$ and $25$.
SST-2 is first introduced by GLUE \citep{wang2018glue}, a famous multi-task benchmark and analysis platform for natrual language understanding, which took a subset from the SST and applied a two-way split (positive or negative) on sentence-level labels.
Owing to the fact that the train/validation/test split are aligned between SST and SST-2, we can run gradient-based methods on the either one of them. Note that we are only working with the test split for both data sets, which contains $2210$ and $1821$ sentences respectively.


\paragraph{Language Model}
We use a popular and publicly available Distill-BERT \citep{sanh2019distilbert} model which is fine-tuned on SST-2 \footnote{\url{https://huggingface.co/distilbert-base-uncased-finetuned-sst-2-english} (Apache License 2.0)}. The accuracy of the Distill-BERT model on SST and SST-2 is $86.6\%$ and $92.4\%$ respectively.

\paragraph{Evaluation Metric}
We extract word-level sentiments from the phrase structure tree (PTB) in SST dataset. 
We take an absolute value after centralization to yield the golden label $\mathcal{M}_{gold}(X)$.
Pearson correlation coefficient, $\rho$ , is the evaluation metric for feature attribution $\mathcal{M}_{gold}(X)$ and $\mathcal{M}_{\cdot}(X)$. The correlation $\rho$ takes value from the range from $-1$ to $1$, and a higher $\rho$ means better feature attribution result.

\paragraph{Main Results}
The main results of the correlation are summarized in Table \ref{table:use_case_2_main}. 
Popular feature attribution methods are listed and compared.
To leverage the problem that some words can have opposite meaning when their sentiment are different, we only limited the sentences neighbor for same category. 
Based on the preliminary experiment, we choose the second layer with $10$ neighbors and $0.39$ cut-off rate, more details about preliminary experiment can be found in Appendix \ref{appendix:B} that using all layers of DistilBERT as the encoder will improve the performance. 

From the result Table \ref{table:use_case_2_main}, it is interesting to point out that the DistilBERT model fine-tuned on SST-2 does not perform equally well on the remaining sentences in SST, so the explanation we yield also has lower correlation for all methods compared with the SST-2. 
Some example and analysis when LAFA works and fails in this dataset by showing neighbor sentences can be found in Appendix \ref{appendix:A}.

We can find out that the InputGrad method outperform SimpleGrad on SST/SST2 as well.
SmoothGrad method achieves a good result by introducing random noise.
From our observation, Sharpley-Value based methods, ``SHAP + Zero'' and ``SHAP + Ref.'' can identify important features with a good chance but may include several irrelevant tokens leading to higher variance. 
Our method LAFA achieves a superior performance in larger average correlation and smaller variance on both SST-2 and SST data sets.




\subsection{Case III: Feature Attribution on Regression Model}


\paragraph{Motivation} 
Amazon's online stores contain rich information about millions of products in product title, brand and description. We hope to better understand the trendy features that affecting price directly from such unstructured raw data, without the need for human labelers / data cleaning. In this application, we fine-tuned a deep language model with price as labels and aim to understand important factors from product descriptions with the given language model.

\paragraph{Data}
We collected the product catalog data of 
about one million
products in personal computer category on Amazon's online store. We concatenate product's title, brand, bullet points and description as the input $X$, and use product price as the label $Y$.

\paragraph{Language Model}
We use BERT-base model and fine-tuned on collected catalog data for price regression.

\paragraph{Evaluation Metric}

To evaluate the performance of feature attribution methods without golden labels, we follow a similar idea as in work \citep{shrikumar2017learning, lundberg2017unified} where the difference of prediction log-odds are measured by deleting pixels with highest importance scores.
In our application, we first randomly select 200 input texts within a threshold of $1\%$ prediction error as evaluation set. 
For each input text, we then mask p\% of the tokens with highest feature attribution scores according to different feature attribution methods.
Then we obtain new prediction result from the masked text denoted as $\hat{y}_{masked}$ and calculate the new mean absolute percentage error (MAPE). 
Higher value of MAPE means that the corresponding method excels in picking important features. 




\paragraph{Main Results}

\begin{figure}[ht]

\resizebox{1\columnwidth}{!}{%
 \centering
  \includegraphics[width=1.1\textwidth]{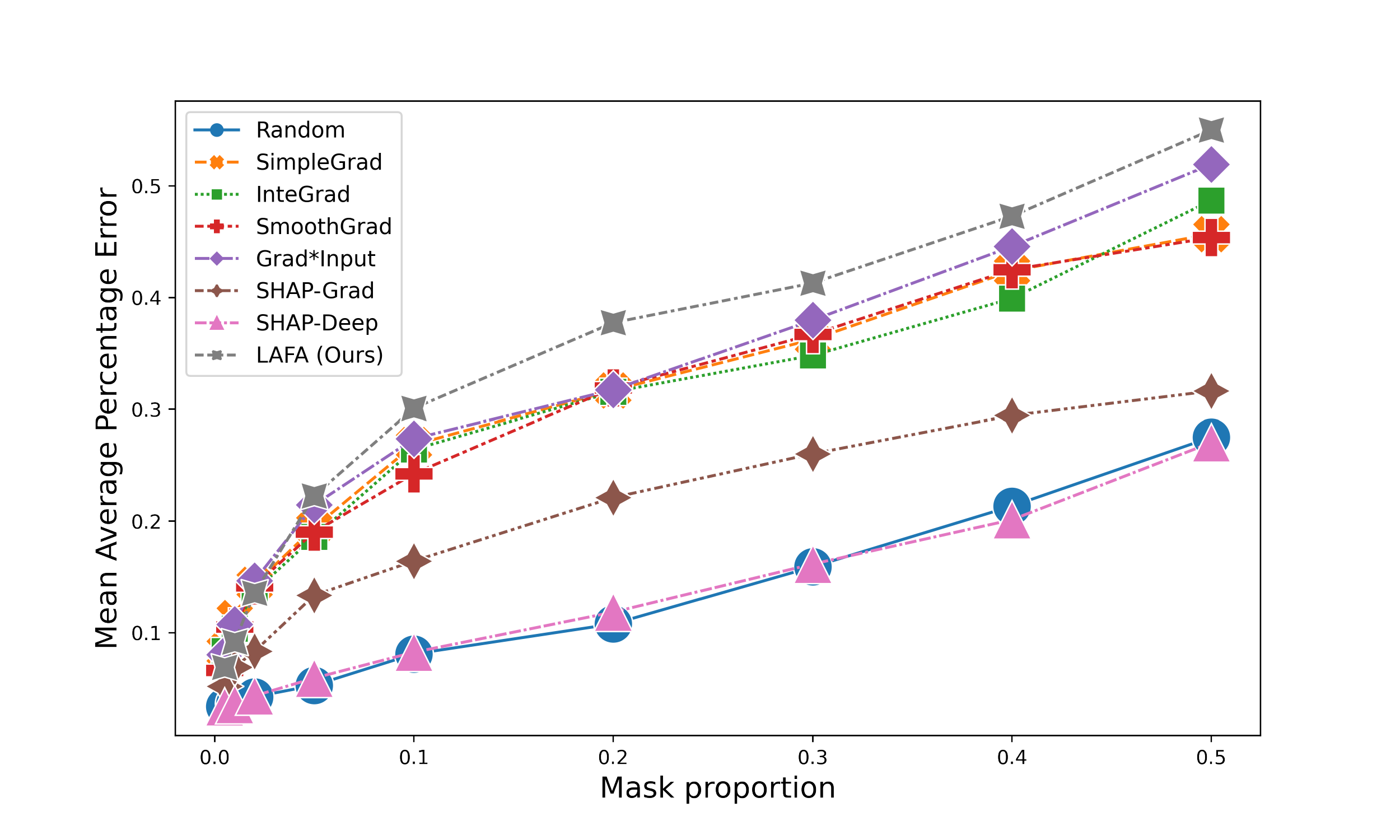}
  }
 \caption{Feature attribution result on Case III. Comparisons of MAPE under different mask proportion. 
}
 \label{fig:ablation_result}
\end{figure}

The results are shown in Figure \ref{fig:ablation_result}  where x-axis is the mask proportion $p$, and y-axis is MAPE.
We observe that the random method has very low MAPE, because randomly masking the input texts will not affect the predicted result as much as the other feature attribution methods.
ShapDeep and ShapGrad also have low MAPE values since simply taking average as aggregation is meaningless in NLP tasks.
Other competing methods have similar performances on this case study and non of these performs better than others in a wide range of mask ratio. 
The proposed LAFA method outperforms other methods by significant margin with masking proportion from $5\%$ to $50\%$, which demonstrates that smoothing over context-level neighbors helps to highlight the important features in similar type of products.

\section{Conclusion}

This paper presents a novel locally aggregated feature attribution method in NLP, which efficiently captures the important features by leveraging similar input texts in the embedding space. We focused on feature attribution of single input based on a fine-tuned model instead of training a language model, henceforth the computation time is of less concern.

One limitation of the LAFA model is that it requires informative neighbor sentences that carry similar information. 
Otherwise, aggregating information from other sentences could be misleading.
Experiments in our datasets show that our method is effective, but the improvements gained from the LAFA varies among different datasets based on the information that neighbor carries.

There are several future directions worthy of study. 
Firstly, labeling feature attribution result in the NLP requires massive human labor, and few datasets are available with golden feature attribution label. Developing new evaluation techniques to further measure model performance is interesting to investigate.
Also, readable feature attribution results could help human beings to develop more business applications. 
For example, developing a key-value pair like \textit{processor}-\textit{i5} as important feature can provide a more plausible feature attribution result to customers.

\bibliography{anthology,custom}
\clearpage
\appendix
\noindent \Large{\textbf{Appendix:}}

\normalsize

\setcounter{table}{0}
\renewcommand{\thetable}{A\arabic{table}}

\section{Model Implementation Detail}
All experiments are conducted with eight NVIDIA Tesla V100 GPUs with 2.5 GHz (base) and 3.1 GHz (sustained all-core turbo) Intel Xeon 8175M processors.

\vspace{-8pt}

\paragraph{Case I} For LAFA, we adopt the cosine function as the kernel function and hyper-parameter with $\lambda = 1$ and SimpleGrad is implemented and aggregated by $\mathcal{M}(x)$ in Equation \eqref{eq:smooth_gradient_our}.

\vspace{-8pt}
\paragraph{Case II} For LAFA, we adopt the Polynomial function as the kernel function $k(\cdot, \cdot) = \mathbb{I}(\cdot, \cdot)$ and hyper-parameter with $\lambda = 0.44$ and SmoothGrad is chosen and aggregated by $\mathcal{M}(x)$ in Equation \eqref{eq:smooth_gradient_our}.

For gradient-based model with hyper-parameters, we tuned them on the first $100$ sentences in the test set. From the grid $[10, 25, 50]$, we choose $25$ as the integral iteration and the smooth candidates.

\vspace{-8pt}
\paragraph{Case III} The indicator function as the kernel function $k(\cdot, \cdot) = \mathbb{I}(\cdot, \cdot)$ with $\lambda = 1$ as hyper-parameter is adopted for LAFA. Neighbor information is aggregated by $\mathcal{M}(x)$ in Equation \eqref{eq:smooth_gradient_our} from the SimpleGrad.

\section{Example of Neighbor Sentences found by Case Studies}\label{appendix:A}
\subsection{Relation Extraction}
\begin{figure}[H]
  \centering
  \includegraphics[width=0.45\textwidth]{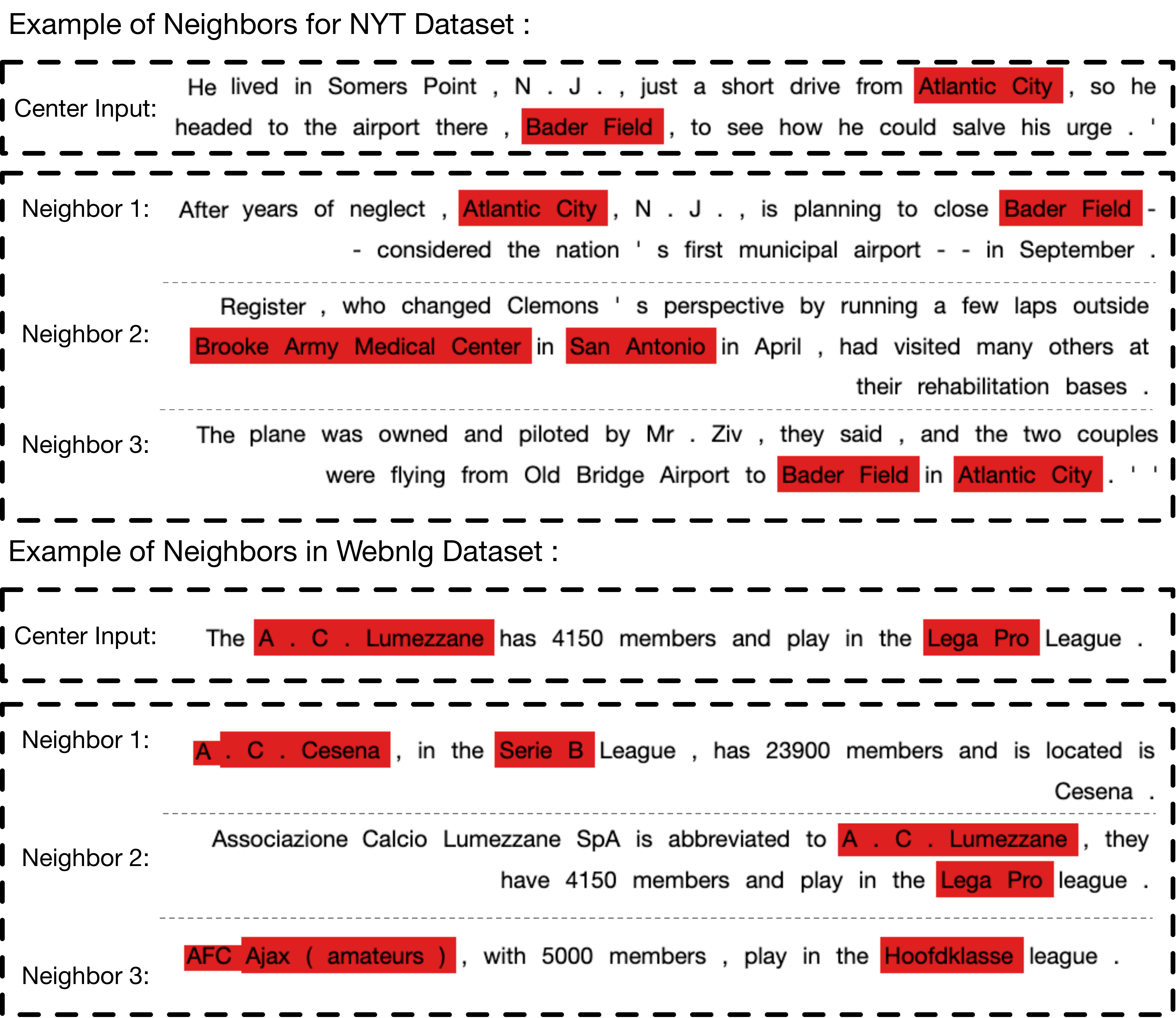}
 \caption{Example of neighbors for NYT and Webnlg. The head and tail entities are highlighted with red color.}
 \label{fig:case1_neighbor}
\end{figure}

In Figure \ref{fig:case1_neighbor}, we show two examples in NYT and Webnlg with their neighbors. We can observe that detected neighbor sentences have a similar meaning, which can be utilized as a reference to help extract the key features from the original sentence.

\subsection{Sentiment Analysis}
\begin{figure}[H]
  \centering
  \includegraphics[width=0.45\textwidth]{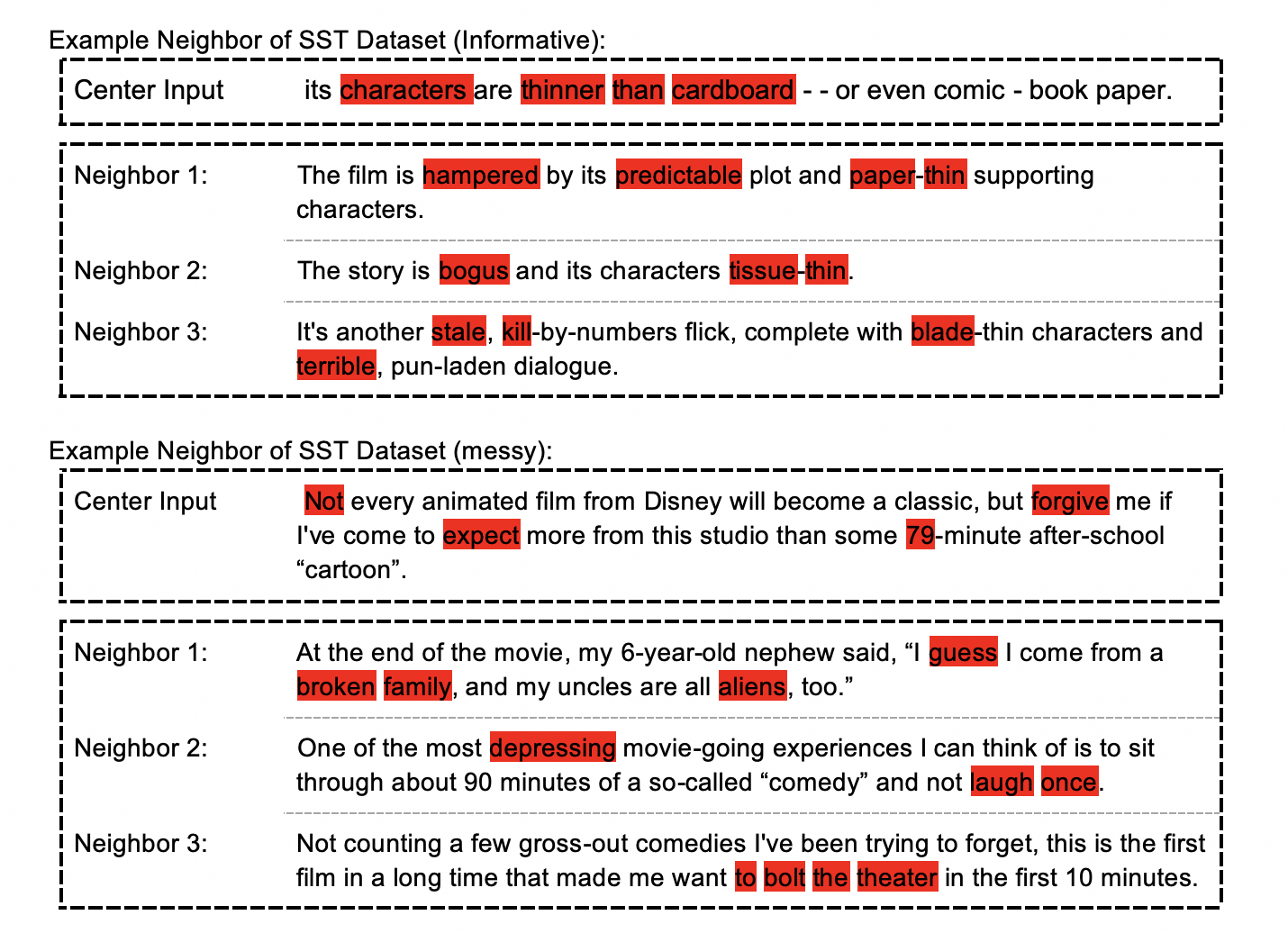}
 \caption{Example of neighbors for the SST data set. Sentiment factors found by SimpleGrad are highlighted with red color.}
 \label{fig:case2_neighbor}
\end{figure}

In SST-2, finding informative neighbors for every sentence is difficult because top sentences may not contain similar tokens, thus does not help. For this reason, we used a cut-off value for this data set to filter out non-informative sentences. In figure \ref{fig:case2_neighbor} we can find two examples from SST, one with ``informative'' good neighbors but another without them. Here for the word ``informative'' we use a quote because we are judging them based on our human understanding.

\section{Examples of Different Feature Attribution Methods under Multiple Cases}
\begin{figure}[H]
  \centering
  \includegraphics[width=0.45\textwidth]{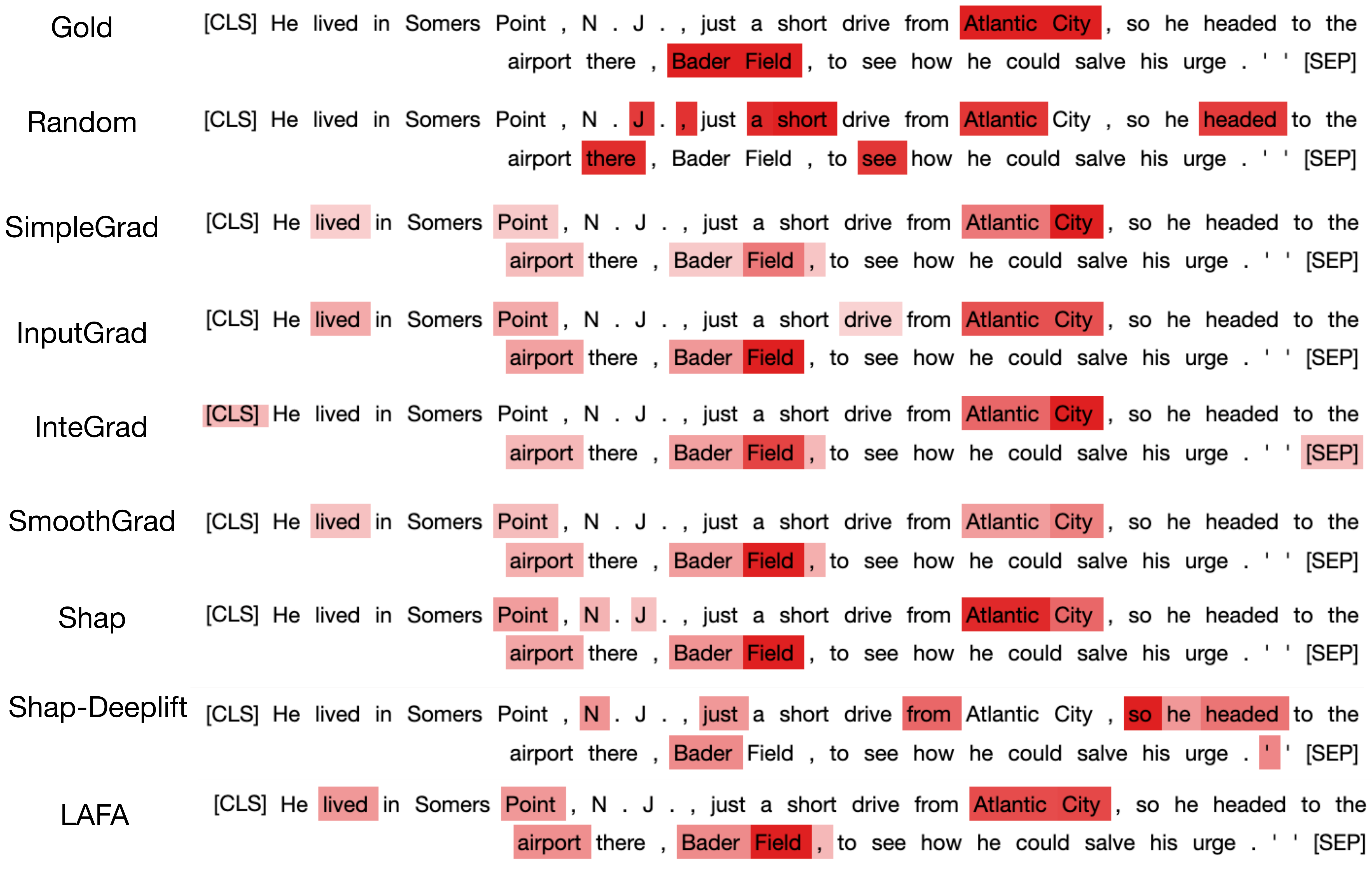}
 \caption{Examples of Case Study I. Important factors are highlighted with red color.}
 \label{fig:case1_exp}
\end{figure}

Here we provide an example in Cases I and II.
In Figure \ref{fig:case1_exp}, LAFA identified locations and ``lived''  as the important factors for relation extraction, and the importance of the ``Atlantic City'' and ``Bader Field'' is stronger than the backbone SimGrad because of aggregation.

\section{Experiment on Different Layer as Neighbor Encoder}\label{appendix:B}
Denote the size of $\mathcal{X}_{sim}$ as $M$, the choice of which can be a critical and challenging task. 
Intuitively, an overly small $M$ would lead to under-smoothing because the target text cannot incorporate enough information from the neighbors.
On the contrary, an overly large $M$ would cause over-smoothing by introducing too much noise.

\begin{figure}[H]
  \centering
  \includegraphics[width=0.45\textwidth]{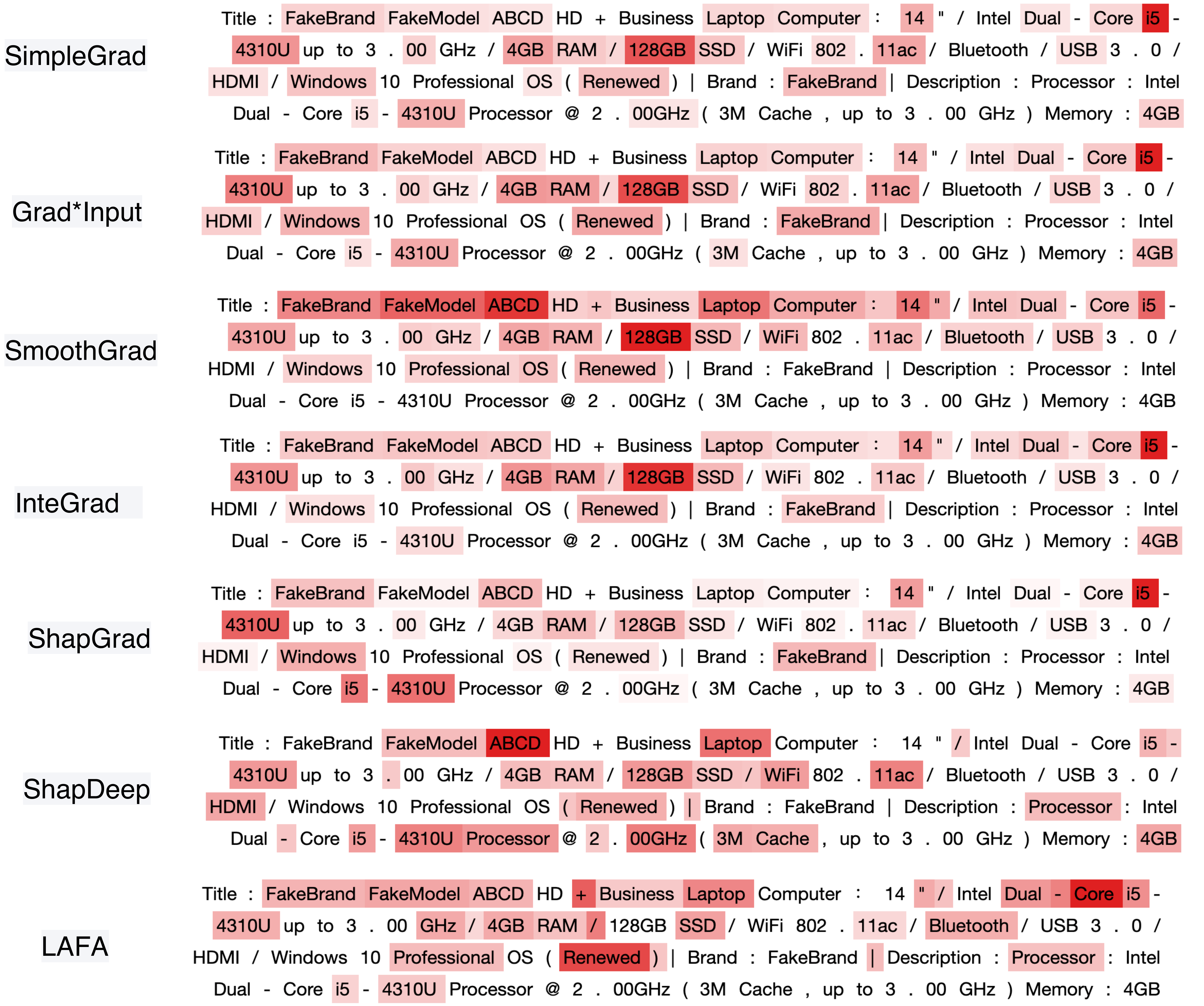}
 \caption{Examples of Case Study III. Important features are highlighted with red color.}
 \label{fig:case2_exp}
\end{figure}

To clarify the neighbor searching process and the difference in the result using different layers, we show some experiments below.

Admittedly, we can directly use the WordPiece embedding as the encoder, which is the input of BERT-based models and enable us to find neighbors in the sense of ``Word Similarity''. However, since the same word can have different meanings in different sentences, and thus different importance in yielded gradients, we might need to use another layer in the BERT model as the encoder to incorporate contextual information. 

We separate the layer search process into two cases depending on the availability of a set of labels that categorizes similar contents into the same group. Generally speaking, both cases recommend the middle layer as the encoder based on our experience.

\subsection{When extra labels are not available}
In the case of SST data, we do not have anything to group similar sentences, so we need to try for different possible layers and find the one that performs the best. 

Here we fixed the max number of neighbors as $10$ and uses $0.05$ quantile of sampled similarities as the cut-off rate to filter those neighbors that are not ``actually close''. We use the SimpleGrad and the SmoothGrad as the baseline for comparison on the first $100$ sentences in the test set.

From table \ref{table:app_sst_3} we can find out that LAFA is a generally good method that always beats the baseline when we use the smooth gradient as the basement method. Layer $2$ performs the best among candidates. The combination of SmoothGrad and Layer $w$ is the final choice and we showed the results on entire SST in the main result part. From here we can find out that for all seven layers, information from faithful neighbors can bring some useful information to an existing sentence.

\begin{table}[H]
\resizebox{\columnwidth}{!}{
\begin{tabular}{lc|lc}
\hline
Method                               & SST\_first100                          & Method          & SST\_first100 \\ 
\hline
SimpleGrad                           & 0.457(0.074)                           & SmoothGrad      &  0.481(0.064)             \\
SpG + LAFA + L1                      & 0.457(0.073)                           & SmG + LAFA + L1 &  0.488(0.063)             \\
SpG + LAFA + L2                      & \textbf{0.458(0.072)}                  & SmG + LAFA + L2 &  \textbf{0.490(0.063)}             \\
SpG + LAFA + L3                      & 0.456(0.072)                           & SmG + LAFA + L3 &  0.489(0.065)             \\
SpG + LAFA + L4                      & 0.456(0.072)                           & SmG + LAFA + L4 &  0.478(0.064)             \\
SpG + LAFA + L5                      & 0.454(0.072)                           & SmG + LAFA + L5 &  0.479(0.063)             \\
SpG + LAFA + L6                      & 0.454(0.073)                           & SmG + LAFA + L6 &  0.483(0.065)             \\
SpG + LAFA + L7                      & 0.456(0.074)                           & SmG + LAFA + L7 &  0.481(0.059)             \\
\hline
\end{tabular}
}
\caption{Feature Attribution Result on first 100 test cases in SST, using simple and smooth gradient as baselines}
\label{table:app_sst_3}
\end{table}

\subsection{When we have extra label}
 \begin{figure}
   \centering
   \includegraphics[width=0.5\textwidth]{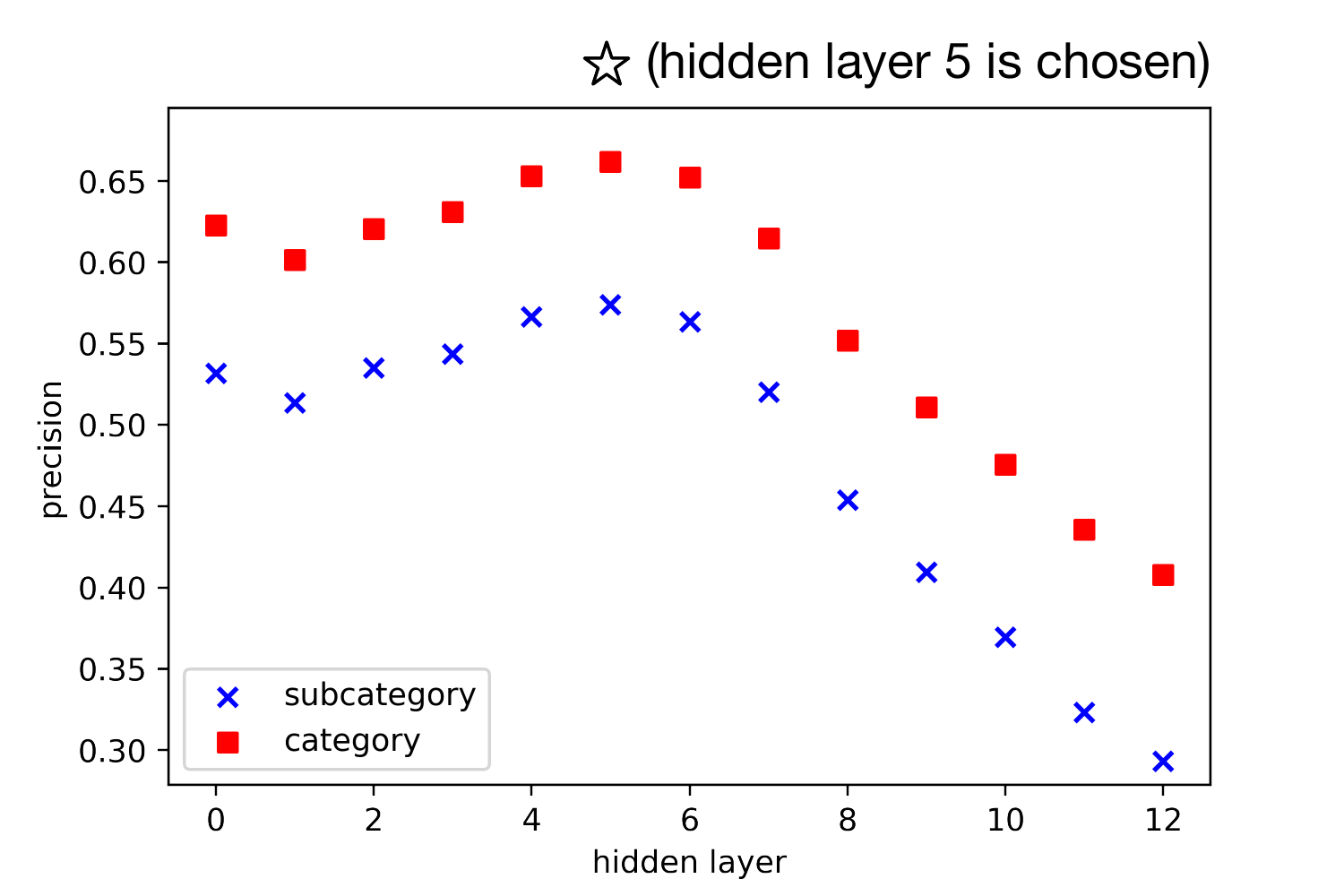}
   \caption{Precision Result in Case Study III}
  \label{fig:encoder_selection}
 \end{figure}
 \vspace{-5pt}
The performance of encoders can be evaluated by the similarity between text $X_0$ and similar texts $\mathcal{X}_{sim}$ obtained from Equation \eqref{eq:similar_text}.
 In this application, we use the product category or subcategory 
 
 \begin{figure*}
  \centering
  \includegraphics[width=0.8\textwidth]{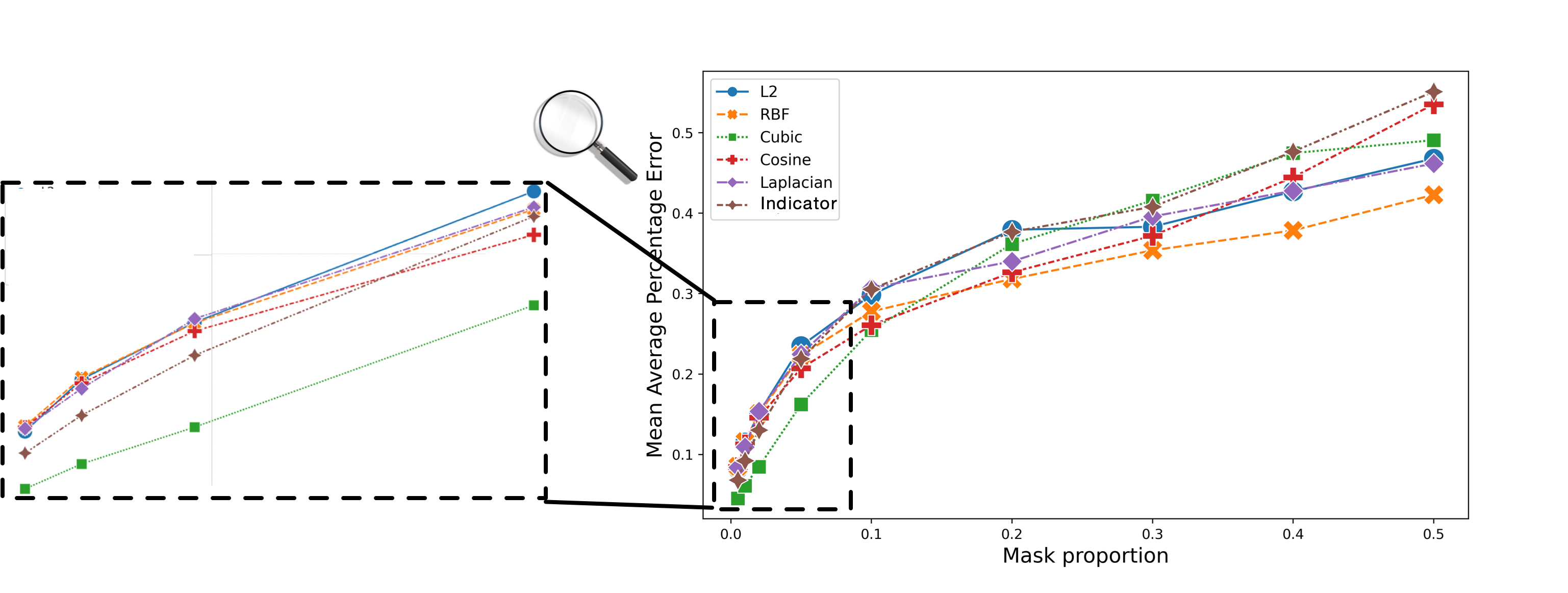}
 \caption{Comparisons for different kernel functions in Case III.}
 \label{fig:case_study_kernel_ablation}
\end{figure*}

\noindent  which is an additional source of labels produced by Amazon to construct a proxy metric to evaluate the similarity. Define the metric of precision as:  
 \begin{equation}\label{eq:prec}
     Precision = \frac{1}{M}\sum_{j = 1}^M \mathbb{I}(c(X_j) = c(X_0)),
 \end{equation}
 where $c(\cdot)$ denotes the category or subcategory of the corresponding product, $\mathbb{I}(\cdot)$ is the indicator function. A high precision represents that the text found $\mathcal{X}_{sim}$ are similar to the text of interest $X_0$. 

 In the numerical study, we randomly sample $10,000$ inputs texts and obtain their corresponding neighbor texts from Equation \eqref{eq:similar_text} with $M = 10$ using each of the $12$ hidden layers in BERT as the encoder $\mathcal{H}_{encoder}$ under $L_2$ norm.
 Figure \ref{fig:encoder_selection} shows the precision result from different encoders, where we observe that the fifth hidden layer has the highest precision in terms of both category and subcategory, which is consistent with the intuition that the middle layer is a trade-off of token-alike and output-alike inputs. In the following experiment, we adopt the fifth layer as the encoder. In general, when no external labels are provided, we may choose a different encoder depending on the use case.

\section{Ablation Study on Kernel Function}\label{appendix:C}

In Case III, we conduct an ablation study with different choices of kernel functions using different mask ratio to find out if different kernel yields different learning speed:

\begin{enumerate}
    \item Radial basis function kernel (\textit{RBF}) :
    \vspace{-3pt}
    \begin{equation*}
        k_{RBF}(a,b) = exp(-||a - b||_2 / l ^ 2),
    \end{equation*}
    \vspace{-3pt}
    where larger hyper-parameter $l$ indicates lower impact from neighbors and vice versa. In the numerical study, we choose $l = 2$ based on the range of embedding $a$ and $b$.
    \item Cubic kernel (\textit{Cubic}):
    \vspace{-3pt}
    \begin{equation*}
        k_{Cubic}(a,b) = (\gamma a^T b + c_0)^d,
    \end{equation*}
    \vspace{-3pt}
    where $\gamma = 7$, $c_0 = 0$ and $d = 3$, smaller $\gamma$ means lower impact from neighbors.
    \item Cosine kernel  (\textit{Cosine}):
    \vspace{-3pt}
    \begin{equation*}
        k_{Cos}(a,b) = a^T b/ ||a|||b|||
    \end{equation*}
    \vspace{-3pt} This kernel function havee no parameter.
    \item Laplacian kernel (\textit{Laplacian}):
    \vspace{-3pt}
    \begin{equation*}
        k_{Laplacian}(a,b) = exp(-||a - b||_1 / l ^ 2),
    \end{equation*}
    \vspace{-3pt}
    in the numerical study, we choose $l = 2$.
    \item L2 norm based similarity (\textit{L2}):
    \vspace{-3pt}
    \begin{equation*}
        k_{L2}(a,b) = 1/clip(||a - b||_2, \lambda_{left}, \lambda_{right}),
    \end{equation*}
    \vspace{-3pt}
    where $clip(\cdot, \lambda_{left}, \lambda_{right})$ denotes clip function with $\lambda_{left} = 0.3$ and $\lambda_{right} = 3$ as clip boundary in numerical study.
    \item Indicator function based similarity (\textit{Indicator}):
    \vspace{-3pt}
    \begin{equation*}
        k_{Indicator}(a,b) = \mathbb{I}(a, b),
    \end{equation*}
    \vspace{-3pt}
    where $\mathbb{I}(\cdot, \cdot)$ denotes indicator function.
\end{enumerate}

The results are shown in Figure \ref{fig:case_study_kernel_ablation}.
We observe that no single kernel function outperforms all other kernel functions under all mask ratios in this study.
Indicator function shows a good performance when the masked ratio is greater than $10\%$, while RBF kernel shows a good performance when the masked ratio is smaller than $5\%$. This can due to the reason that the indicator function only aggregates identical words and this conservative manner helps when we lost most important words. 

\end{document}